\documentclass[11pt]{article}

\PassOptionsToPackage{table}{xcolor}
\usepackage[final]{acl}
\usepackage{times}
\usepackage{latexsym}
\usepackage[T1]{fontenc}
\usepackage[utf8]{inputenc}
\usepackage{microtype}
\usepackage{inconsolata}
\usepackage{graphicx}
\usepackage{booktabs}
\usepackage{array}
\usepackage{tabularx}
\usepackage{amsmath,amssymb}
\usepackage{xcolor}
\usepackage{url}
\usepackage{listings}
\usepackage[most]{tcolorbox}
\usepackage{placeins}
\usepackage{float}
\usepackage{needspace}
\usepackage{stfloats}

\definecolor{TableHeaderShade}{RGB}{235,237,232}
\definecolor{MethodBlue}{RGB}{232,241,250}
\definecolor{TableGroupShade}{RGB}{245,247,243}
\definecolor{promptbg}{HTML}{F7F8FA}
\definecolor{promptframe}{HTML}{8A8F94}
\definecolor{prompttitle}{HTML}{E8EDF1}
\lstdefinestyle{paperprompt}{
  basicstyle=\ttfamily\fontsize{8.8pt}{8.9pt}\selectfont,
  breaklines=true,
  breakatwhitespace=false,
  columns=fullflexible,
  keepspaces=true,
  showstringspaces=false,
  upquote=true,
  tabsize=2
}
\newtcblisting{promptbox}[1]{
  enhanced,
  breakable,
  title after break={Continued},
  extras title after break={fonttitle={\bfseries\footnotesize}},
  listing only,
  title={#1},
  colback=promptbg,
  colframe=promptframe,
  colbacktitle=prompttitle,
  coltitle=black,
  fonttitle={\bfseries\footnotesize\raggedright\hyphenpenalty=10000\exhyphenpenalty=10000},
  boxrule=0.55pt,
  arc=2pt,
  boxsep=1pt,
  left=4pt,
  right=4pt,
  top=1pt,
  bottom=1pt,
  before skip=1pt,
  after skip=1.5pt,
  listing options={style=paperprompt}
}

\newcommand{\ogrank}{\textsc{OG-Rank}}
\title{One Score, Two Speeds: Selective Generation\\for Clinical Order Entry}

\author{
  Praphul Singh \quad
  Akshat Agarwal \quad
  Sri Gadde \quad
  Krishnaram Kenthapadi \\
  Oracle Health AI
}

\begin{document}
\maketitle

\begin{figure*}[b]
  \centering
  \includegraphics[width=0.99\textwidth,trim=0 18.75bp 0 31.5bp,clip]{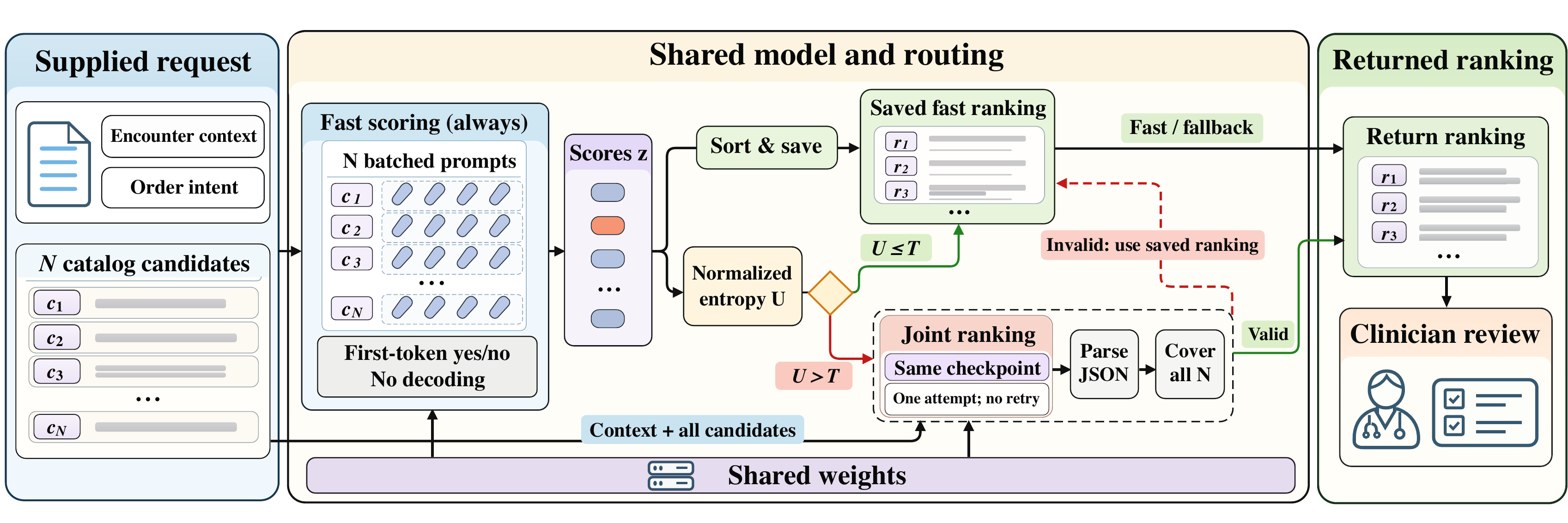}
  \caption{One checkpoint supplies fast scores, a saved ranking, and entropy $U$. If $U>T$, it generates a joint ranking; invalid output returns the saved ranking without retry. Experiments use $N=20$ candidates and $T=0.90$.}
  \label{fig:system}
\end{figure*}

\begin{abstract}
Clinical order catalogs often contain several entries for the same medication, test, or procedure, distinguished by details such as dose, route, or imaging protocol. Selecting among them requires both the clinician's ordering intent and encounter context. A language model can compare these candidates by generating a complete ranking, but doing so for every request adds latency. We introduce \ogrank{}, a two-speed reranker that uses one adapted model for fast candidate scoring and selective joint-ranking generation. The same scores produce the fast ordering and determine whether to spend additional computation on the list, avoiding a separate routing model. On 214 de-identified ordering cases, selective generation raises top-rank agreement with clinician-signed orders from 44.9\% to 56.1\%, retaining 70.6\% of the net gain from always generating while avoiding generation on 55\% of requests. Mean reranking latency falls from 5.0 to 2.7\,s. Matched-budget comparisons favor entropy routing over a top-two score margin on three same-workflow sets. The results support score reuse as a practical way to allocate joint comparison within a clinician-reviewed order-entry workflow; the evaluation measures historical-order agreement conditional on candidate availability.
\end{abstract}

\section{Introduction}
\label{sec:introduction}

When a clinician intends to order a medication, laboratory test, imaging exam, or procedure, the electronic health record (EHR) represents that intent as an entry in the institution's order catalog.
Computerized order entry is consequential to clinical workflow \cite{bates1998cpoe}, while data-driven systems can anticipate orders clinicians may need \cite{chen2015orderrex,wang2020clinicnet}.
Our problem begins after intent expression and candidate retrieval.
An upstream system narrows the catalog to 20 candidates, but several may express the same clinical concept while differing in dose, route, anatomy, modality, or protocol; 54\% of primary test lists contain near-duplicate names.
The reranker places the entry best matching the clinician's intent and encounter context first.
We call the clinician's finalized selection the \emph{signed-order target}; agreement with it measures workflow agreement, not clinical correctness.

This setting creates a quality--latency tradeoff.
Bi-encoders encode queries and candidates separately, while cross-encoders jointly score each query-candidate pair \cite{reimers2019sbert,nogueira2019passage}; decoder rerankers can instead compare a supplied candidate list by generating its complete ordering \cite{sun2023rankgpt,ma2023lrl}.
Candidate-wise scoring is predictable and batchable.
Joint-ranking generation presents all 20 candidates together, enabling cross-candidate comparison but requiring sequential decoding and output validation.
We call this complete-ranking call the \emph{slow path}; \emph{full slow} invokes it for every list.
We ask whether the fast scores can allocate a limited number of these calls to improve top-rank agreement.

\ogrank{} addresses this question with one 3-billion-parameter decoder model (Figure~\ref{fig:system}).
First-position yes/no token scores provide candidate relevance scores and a ranking without decoding text.
Entropy summarizes how concentrated those scores are across the candidate list and controls routing.
We select a threshold once on a 200-case development set containing no evaluation-set patients, under a prespecified slow-path call cap, then freeze it.
Concentrated lists return the fast ranking; diffuse lists permit one slow-path call.
A generated ordering replaces the fast ranking only if it parses and covers every candidate; otherwise, the fast ranking is retained.
The routing comparison tests whether this score concentration helps allocate generation at a fixed call budget.

We evaluate the frozen policy on 214 de-identified ordering cases from 187 patients in an operational EHR workflow.
Evaluation requires the signed-order target in the supplied top-20 list; retrieval naturally includes it in 205 cases, and it replaces the final candidate in the remaining nine without revealing its status.
Recall@1, the fraction of cases with the target ranked first, rises from .449 to .561 with 96 slow-path calls (45\%).
At the same call count, routing by the gap between the two highest fast scores, called the margin, yields .519; full slow yields .607 using 214 calls.
Mean reranking latency is 2.7\,s, compared with 5.0\,s for full slow in the measured environment.
No evaluated output affected care.

\paragraph{Contributions.}
\textbf{(1) One score vector for ranking and routing.}
A single checkpoint reuses first-step scores to produce a complete no-decoding ranking and decide which lists receive one joint-ranking refinement.
\textbf{(2) A controlled fixed-checkpoint routing analysis.}
We hold the checkpoint, candidate lists, scores, and deterministic generated rankings fixed when comparing entropy and margin at the same number of slow-path calls, and quantify their paired differences with patient-clustered uncertainty estimates.
\textbf{(3) Bounded generation with measured fallback.}
The slow path permits one validated generation attempt and retains the fast ranking as fallback; we report reranking latency and request-level validation outcomes.

\section{Related Work and Positioning}
\label{sec:related}

\paragraph{Scoring and generative reranking.}
Prior systems score candidates with bi-encoders and cross-encoders or rank them from the likelihood of generated relevance labels \cite{reimers2019sbert,nogueira2019passage,nogueira2020monot5,zhuang2024beyond}.
Decoder rerankers can instead decode one ordering over all supplied candidates, commonly called listwise generation \cite{sun2023rankgpt,ma2023lrl}; Faster Improved Listwise Reranking with Single Token Decoding (FIRST) ranks from first-position identifier scores without completing a response \cite{reddy2024first}.
\ogrank{} reuses one first-step score vector for ranking, routing, and fallback.

\paragraph{Selective computation.}
Selective prediction, adaptive computation, and model cascades vary whether or how much computation a request receives \cite{geifman2017selective,graves2016adaptive,chen2024frugalgpt}.
CoRanking combines small and large ranking agents, while GroupRank compares candidate groups to balance context across candidates and efficiency \cite{liu2025coranking,long2026grouprank}.
Confidence-Adaptive Thinking (CAT) changes reasoning length, and AcuRank repeats comparisons over uncertain subsets \cite{jiang2026cat,yoon2025acurank}.
At one fixed call count, we compare uniform allocation, which ignores the scores, with margin and entropy routing.

\paragraph{Clinical boundary.}
Prior systems use EHR histories to recommend individual orders or predefined bundles of orders \cite{chen2015orderrex,wang2020clinicnet}, or to predict future events \cite{rajkomar2018scalable}.
Our task begins after an upstream system derives the intended order from encounter context and retrieves candidates; it only reorders supplied local-catalog candidates for clinician review.

\section{Two-Speed Serving Policy}
\label{sec:serving}

Serving must bound generative computation while retaining a valid ranking when generation is skipped or fails.
Given encounter context $x$ and a fixed candidate set $\mathcal{C}=\{c_1,\ldots,c_N\}$, \ogrank{} uses three modes: \emph{fast} sorts scores without decoding text, \emph{two-speed} applies the entropy gate and may invoke slow path, and \emph{full slow} invokes that path for every list.
All modes return supplied candidates for downstream review; routing requires no separate model call.

\paragraph{Candidate-wise scoring without text decoding.}
For each candidate $c_j$, we render a candidate-wise, or pointwise, prompt $u_j$ that presents the encounter context, optional structured patient fields, the highlighted candidate, and compact alternatives, and asks for a yes/no assessment of fit.
Let $\ell_{j,v}$ be the model logit, an unnormalized score, for token $v$ at the first output position.
We predefine disjoint sets $\mathcal{Y}$ and $\mathcal{N}$ of single tokens that the model's tokenizer uses for supported variants of \texttt{yes} and \texttt{no}, respectively.
The fast score is
\begin{equation}
z_j =
\log\!\sum_{v\in\mathcal{Y}}\exp(\ell_{j,v})
-
\log\!\sum_{v\in\mathcal{N}}\exp(\ell_{j,v}).
\label{eq:fast-score}
\end{equation}
The common normalization over the full token vocabulary cancels, so $z_j$ is the log ratio of pooled yes and no probability mass at the first output position.
No token is sampled or decoded; candidates are ranked by decreasing $z_j$.
We batch all $N=20$ prompts within a request. Each prompt produces one candidate score; the slow path instead generates one ordering over the list.

\paragraph{Routing from score concentration.}
The gap between the two highest scores describes competition for first place, but leaves out the rest of the list.
Two lists can have the same leading gap while one contains several other candidates with similar scores and the other has only two competitive candidates.
Entropy distinguishes these cases by considering all candidate weights.
This distinction motivates its use for catalogs containing groups of near-duplicate entries, where several doses, routes, or protocols may remain plausible under the supplied context.
The matched-budget comparison with margin routing tests whether this broader summary helps choose which lists receive joint comparison.

The same scores yield a list-level routing signal:
\begin{align}
\widetilde{p}_j &= \frac{\exp(z_j)}{\sum_{k=1}^{N}\exp(z_k)}, \\
U(x,\mathcal{C}) &=
-\frac{\sum_{j=1}^{N}\widetilde{p}_j\log\widetilde{p}_j}{\log N}.
\label{eq:list-ambiguity}
\end{align}
The normalized candidate weights $\widetilde{p}_j$ sum to one, and their normalized entropy $U\in[0,1]$ is small when one candidate dominates and large when scores are similar.
Computing $U$ requires only normalization and entropy arithmetic over the $N=20$ scalar scores already produced for fast ranking, with no additional model invocation or decoding; its runtime contribution was not isolated separately.
$U$ is a routing statistic, not calibrated clinical uncertainty.
Before test evaluation, we selected and froze $T=0.90$ on a 200-case development set containing no evaluation-set patients by maximizing Recall@1 under a prespecified 45\% cap.
This study cap keeps a majority of development requests in fast mode but is not a production-capacity estimate or universal rate.
If $U\leq T$, the service returns the fast ordering; if $U>T$, it permits at most one bounded slow-path call.
Figure~\ref{fig:policy-evidence} reports neighboring descriptive points; test-set slow-path fractions follow from the frozen threshold and were not retuned. Appendix~\ref{app:threshold-selection} gives the selection rule and resampling analysis.

\paragraph{Bounded joint-ranking generation and fast-path fallback.}
The slow path sends the encounter context and all supplied candidates to the same checkpoint in a separate prompt, using greedy decoding with a 512-token output cap.
It makes one complete-ranking attempt for a JavaScript Object Notation (JSON) object containing a candidate ordering and brief rationale.
The service accepts only parseable JSON whose ranking maps to and covers the supplied candidate set; otherwise it returns the fast ordering. The rationale does not affect acceptance.
Let $\pi_f$ be the fast ordering, $g$ the generated object, and $V(g;\mathcal{C})$ the true/false validation check that $g$ has the required structure and exactly covers $\mathcal{C}$. The returned ordering $\pi_{\mathrm{OG}}$ is
\begin{equation}
\pi_{\mathrm{OG}}(x,\mathcal{C}) =
\begin{cases}
\pi_f, & U\leq T,\\
\operatorname{parse}(g), & U>T\ \land\ V(g;\mathcal{C}),\\
\pi_f, & U>T\ \land\ \neg V(g;\mathcal{C}).
\end{cases}
\label{eq:serving-policy}
\end{equation}
No repair or retry is issued; $\operatorname{parse}(g)$ denotes the validated ordering extracted from $g$.
Appendices~\ref{app:serving-interface}--\ref{app:prompts} specify the serving control flow, reproducibility boundary, and representative prompt contracts.

\section{Model Adaptation}
\label{sec:training}

To make one checkpoint serve both paths, we adapt Llama-3.2-3B-Instruct \cite{meta2024llama32} with low-rank adaptation (LoRA) \cite{hu2022lora} on approximately 10,000 pointwise prompts from roughly 500 encounters.
Supervised adaptation precedes Group Relative Policy Optimization (GRPO)-style updates \cite{liu2024deepseekmath} using a fixed automated rubric over decision agreement, clinical relevance, specificity, safety, and format.
The automated judge applying this rubric is confined to offline reward computation: it supplies neither test labels nor serving decisions, and its rubric dimension named \emph{safety} is not a clinical safety assessment.
At serving time, the same weights produce first-step yes/no scores for fast ranking and answer a separate prompt containing all supplied candidates for routed refinement.
The slow path therefore reuses pointwise-trained weights without direct supervision on complete candidate orderings.
Appendix~\ref{app:training-details} details example construction, optimization, rewards, difficulty-based allocation, and this complete-ranking reuse.

The successive checkpoints improve the fast ranking before selective generation is introduced.
On the primary set, fast-mode Recall@1 is .25 for the public checkpoint, .40 after supervised pointwise adaptation, and .449 after rubric-guided updates (Table~\ref{tab:training-ablation}).
These measurements follow one training sequence, with each stage building on the previous checkpoint.
The routing experiments then keep the final adapted weights fixed: fast-only inference scores .449, selective generation .561, and full generation .607.
This separates two questions in the evaluation: how the fast ranking changes during adaptation, and how much a serving policy can gain by allocating generation once adaptation is complete.
The entropy--margin comparison addresses the latter question at a common call budget.

\section{Evaluation Design}
\label{sec:evaluation}

\paragraph{Task inputs and evaluation labels.}
We evaluate 214 de-identified ordering cases from 187 patients in an operational EHR workflow. Each case contains encounter context, available patient information, and a clinician-signed finalized order. Evaluation cases were excluded from the approximately 500 encounters used for adaptation. The 200-case development set contains no evaluation-set patients.

An upstream GPT-4o component derives a free-text order intent from encounter context without access to the signed order. The intent retrieves the 20 highest-ranked catalog-eligible candidates. In the primary set, 62\% of intents name a concrete medication, test, or imaging examination, while 38\% express a broader clinical goal. Candidate lists often retain fine distinctions: 54\% contain near-duplicate names, 28.5\% dose or route variants, and 21\% imaging protocol or modality variants; these categories overlap.

We evaluate reranking conditional on candidate availability. The signed-order target occurs naturally in 205 of 214 retrieved lists; in the remaining nine, it replaces the final candidate without a target marker. Every policy receives the same list. Recall@1 measures the fraction of lists with the signed-order target first; normalized discounted cumulative gain through rank 20 (nDCG@20) also credits upward movement below the first position. Both measure agreement with historical clinician selections. Appendix~\ref{app:data-profile} gives the construction and metrics, and Section~\ref{sec:results} checks sensitivity to the nine replacements.

\begin{table*}[t]
\centering
\small
\renewcommand{\arraystretch}{1.12}
\begin{minipage}[t]{0.56\textwidth}
\textbf{(a) Primary set: 214 lists}
\par\vspace{3pt}
\setlength{\tabcolsep}{3pt}
\begin{tabularx}{\linewidth}{@{}>{\raggedright\arraybackslash}Xrrr@{}}
\toprule
\rowcolor{TableHeaderShade}
\textbf{Policy} & \textbf{Slow calls} & \textbf{Matches} & \textbf{Recall@1} \\
\midrule
Fast only & 0 & 96 & .449 \\
Uniform (expected) & 96 & 111.25 & .520 \\
Margin gate & 96 & 111 & .519 \\
\rowcolor{MethodBlue}
\textbf{Two-speed (entropy)} & 96 & 120 & .561 \\
Full slow & 214 & 130 & .607 \\
\bottomrule
\end{tabularx}
\end{minipage}\hfill
\begin{minipage}[t]{0.41\textwidth}
\textbf{(b) Matched-budget gains over fast}
\par\vspace{3pt}
\setlength{\tabcolsep}{3pt}
\begin{tabularx}{\linewidth}{@{}>{\raggedright\arraybackslash}Xrrr@{}}
\toprule
\rowcolor{TableHeaderShade}
\textbf{Set} & \textbf{$n$} & \textbf{Margin} & \textbf{Entropy} \\
\midrule
Primary & 214 & +15 & +24 \\
A & 250 & +16 & +25 \\
B & 300 & +18 & +28 \\
\bottomrule
\end{tabularx}
\par\vspace{4pt}
\raggedright
Primary entropy minus margin:\\
\textbf{+9 matches}, 95\% CI $[+2,+16]$.
\end{minipage}
\caption{\textbf{Entropy allocates the same generation budget more effectively.}
(a) All policies share the checkpoint, candidate lists, fast scores, and fixed slow outputs. Uniform allocation is the expectation over assignments of exactly 96 calls.
(b) Margin is matched to entropy's call count within each set. Gains count additional signed-order matches; the primary interval uses 1,000 paired patient-clustered bootstrap resamples. Higher matches and Recall@1 indicate better historical-order agreement.}
\label{tab:main_results}
\end{table*}

\begin{figure*}[t]
  \centering
  \includegraphics[width=0.99\textwidth]{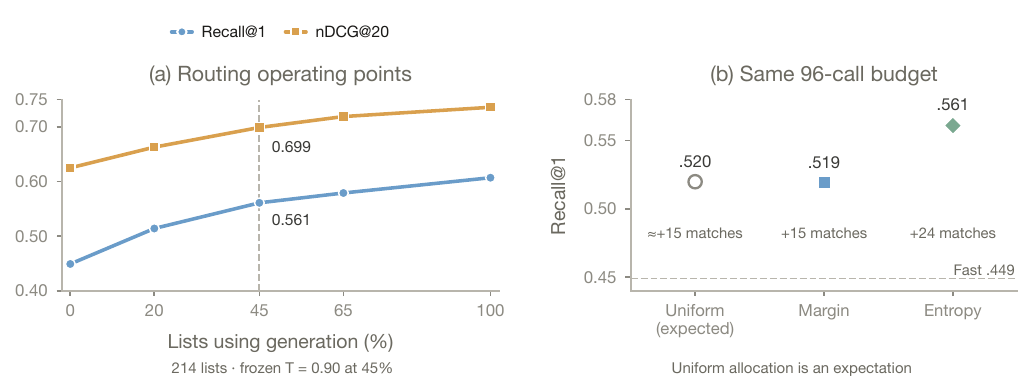}
  \caption{\textbf{Controlled fixed-checkpoint routing and ranking quality versus slow-path use.}
  (a) Observed entropy-routing points from fast to full slow; slow-path fraction is the share of lists receiving a complete-ranking LLM call, and frozen $T=0.90$ is highlighted.
  (b) At one invocation count, both measured gates reuse the same scores and slow outputs: margin routes the smallest top-two gaps, while uniform allocation is the random-assignment expectation. Table~\ref{tab:main_results} reports paired uncertainty for the nine-match difference.}
  \label{fig:policy-evidence}
\end{figure*}

\paragraph{Fixed-budget routing comparison.}
The frozen threshold routes 96/214 primary lists (44.86\%) to generation. To isolate which lists receive that computation, we compare entropy with a margin gate that routes the 96 smallest gaps between the two highest fast scores. Uniform allocation gives the expected outcome of assigning 96 calls without using scores. These comparisons hold the checkpoint, candidates, fast scores, and deterministic slow outputs fixed. Fast-only and full-slow inference define the zero-call and all-call endpoints.

We estimate paired differences with 1,000 patient-clustered bootstrap resamples over the 187 primary-set patients, retaining cases from the same patient together. Appendix~\ref{app:outcome-ledger} reports these intervals alongside the original case-level analysis. We also examine neighboring operating points and development-set threshold selection under 40\% and 50\% call caps.

\paragraph{Encoder references.}
The MS-MARCO MiniLM-L12-v2 cross-encoder was fine-tuned on the same clinical training data as \ogrank{} and achieves .18 Recall@1 and .437 nDCG@20 \cite{sentenceTransformers2026minilm}. The PubMedBERT bi-encoder reaches .13 and .364 \cite{gu2021pubmedbert}. These results contextualize ranking performance; the allocation comparison above uses a single adapted checkpoint so that changing the model does not confound the routing effect.

\paragraph{Additional sets.}
We apply the frozen configuration to 250-case Additional Set A and 300-case Additional Set B from the same workflow to test whether the ranking gains recur on further cases.
Appendix~\ref{app:holdouts} reports the complete metrics and scope.

\paragraph{Serving measurements.}
Timing uses one NVIDIA A100 80\,GB GPU, CUDA 12.2, and vLLM 0.6.1. Request concurrency is one, with the 20 fast-scoring prompts batched within each request. Slow generation uses greedy decoding, a 512-token output cap, and one attempt without repair or retry. Mean input length is 1,142 tokens, and accepted slow outputs average 187 tokens. The measured interval begins with supplied context and candidates and ends with the returned ranking after validation or fallback. GPT-4o intent derivation and catalog retrieval precede this interval. Table~\ref{tab:latency-training} reports reranker latency; Appendix~\ref{app:serving-interface} documents the configuration and validation outcomes.

\paragraph{Difficulty slices.}
We report paired metrics for two overlapping case subsets, or slices: 96 high-ambiguity cases and 61 medication dose/route cases. These annotations describe candidate-list content and remain fixed across policies. They describe content-based difficulty; entropy routing is defined separately by the model scores.
Appendix~\ref{app:slice-metrics} reports the subset metrics, and Appendix~\ref{app:internal-review} documents a separate 40-case qualitative staff review.

\begin{table}[tb]
\centering
\small
\renewcommand{\arraystretch}{1.12}
\setlength{\tabcolsep}{4pt}
\begin{tabular}{@{}lrrr@{}}
\toprule
\rowcolor{TableHeaderShade}
\textbf{Metric} & \textbf{Fast} & \textbf{Two-speed} & \textbf{Full slow} \\
\midrule
Slow path \% & 0    & 45   & 100 \\
Recall@1      & .449 & .561 & .607 \\
nDCG@20       & .625 & .699 & .736 \\
\midrule
p50 (s)       & 0.4  & 0.6  & 4.9 \\
p95 (s)       & 0.8  & 5.8  & 6.2 \\
Mean (s)      & 0.5  & 2.7  & 5.0 \\
\bottomrule
\end{tabular}

\caption{\textbf{Measured reranking latency.}
Request-level values span the supplied reranking input through validation or fallback for the same 214 requests in one execution environment.}
\label{tab:latency-training}
\end{table}

\begin{figure*}[t]
  \centering
  \includegraphics[width=0.98\textwidth]{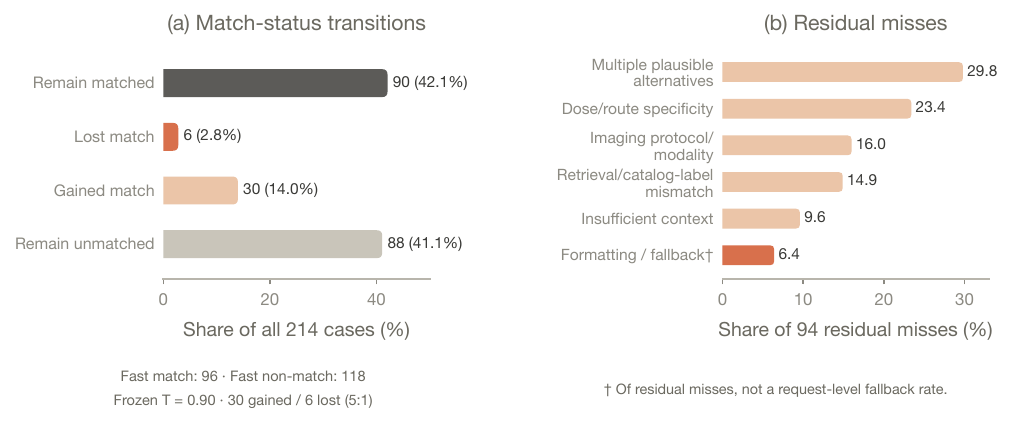}
  \caption{\textbf{How selective ranking calls change historical agreement.}
  (a) Of 36 changed Recall@1 outcomes among 214 lists, 30 gain signed-order agreement and 6 lose it.
  (b) Descriptive categories among 94 residual misses. The label is a historical order, not correctness; formatting/fallback is not a request-level fallback fraction.}
  \label{fig:outcome_analysis}
\end{figure*}

\section{Selective Routing Results}
\label{sec:results}

\paragraph{Ranking quality at a fixed generation budget.}
Fast scoring matches 96 of 214 signed-order targets. Entropy routing increases this to 120 using 96 generation calls, while full slow reaches 130 using 214 calls (Table~\ref{tab:main_results}). At the same 96-call budget, the margin gate obtains 111 matches. Entropy therefore adds nine matches over margin, with a 95\% patient-clustered bootstrap interval of $[+2,+16]$. Uniform allocation has an expectation of 111.25 matches. The comparison supports the use of the full score distribution to allocate generation in this workload.

The advantage recurs on both additional sets. On Set A ($n=250$), entropy adds 25 matches over fast and margin adds 16. On Set B ($n=300$), their gains are 28 and 18. Entropy's advantage over margin is therefore nine, nine, and ten matches across the three sets. These additional comparisons use the same workflow and matched generation budgets within each set.

\paragraph{What changes when generation is invoked?}
Among the 96 routed primary lists, generation creates 30 signed-order matches and loses six; 90 cases remain matches and 88 remain misses over the full set. The net gain of 24 has a 95\% patient-clustered bootstrap interval of $[+15,+32]$. Selective routing thus retains $24/34=70.6\%$ of full generation's net gain while invoking generation for 44.9\% of requests. The six losses matter: a diffuse fast-score distribution can justify additional comparison without guaranteeing that generation improves an individual ranking. Figure~\ref{fig:outcome_analysis} gives the paired outcomes and residual-error categories.

\paragraph{Budget and candidate-construction sensitivity.}
Recall@1 rises from .449 through .514, .561, and .579 to .607 at the displayed 0\%, 20\%, 45\%, 65\%, and 100\% generation points (Figure~\ref{fig:policy-evidence}). Selecting thresholds on development data under 40\% and 50\% call caps changes primary Recall@1 by at most .014 and preserves the router ordering. Excluding the nine lists requiring target replacement changes each of fast, entropy, and full-slow Recall@1 by at most .009. Entropy's gain over fast remains 23 matches, compared with 24 on all lists. The gain therefore persists among the 205 lists where retrieval naturally includes the target.

\paragraph{Latency and fallback reliability.}
Mean reranking latency falls from 5.0\,s for full slow to 2.7\,s for entropy routing. Its median is 0.6\,s, close to fast-only inference at 0.4\,s, while p95 remains 5.8\,s versus 6.2\,s for full slow (Table~\ref{tab:latency-training}). Selective routing primarily reduces how often requests incur generation, leaving a slow tail for those that do.

Of the 96 routed requests, 91 (94.8\%) yield valid JSON rankings covering every candidate. Three fail parsing and two return incomplete rankings. All five rejections return the saved fast ranking without retry. Validation success is 93--96\% on Sets A and B. Appendix~\ref{app:serving-interface} distinguishes these request-level counts from residual ranking-error categories.

\paragraph{Specification-sensitive cases.}
On the annotated high-ambiguity subset, Recall@1 rises from .36 to .52; on the dose/route subset, it rises from .43 to .57. These overlapping slices describe where agreement improves, while the matched-budget comparison tests how computation is allocated. Multiple plausible alternatives and dose/route specificity account for 29.8\% and 23.4\% of the 94 residual misses, respectively. Appendices~\ref{app:slice-metrics} and~\ref{app:outcome-ledger} report the full descriptive breakdown.

\section{Discussion}
\label{sec:discussion}

The main benefit of score reuse is that routing adds no separate model or learned confidence estimator. Candidate scoring already provides the information used to allocate joint comparison. Holding the checkpoint, lists, scores, and slow outputs fixed makes the entropy--margin comparison a test of that allocation: the nine additional primary-set matches come from choosing different requests for the same generation budget. The result supports entropy as a useful routing rule in this workload, while the six losses among routed cases show why it should not be treated as a guarantee of individual improvement.

The full-slow result also shows what is gained by extending generation to the requests left on the fast path.
The 96 calls selected by entropy yield 24 net additional matches; generating on the remaining 118 requests adds another ten.
Because both policies use the same slow outputs, this difference follows directly from their paired outcomes.
It corresponds to 25 net additional matches per 100 calls on the selected lists, compared with about 8.5 on the remaining lists.
These rates count gains minus losses across each group.
Generation remains useful outside the selected set, but its lower aggregate return explains how the selective policy retains most of the full-slow gain with fewer than half as many calls.

The latency profile also determines where this design is useful. Most requests avoid decoding and return close to fast-path latency, but requests sent to generation retain a slow tail. A service with a strict per-request deadline would therefore need a different policy from one optimizing mean response time under a compute budget. The measured 45\% operating point is one such budget; moving to 65\% adds four net matches. A new site should select its threshold using local development data and measure performance under its own arrival rate and concurrency. Intent derivation and retrieval add costs outside the measured reranker interval.

Retaining the fast ranking gives generation a bounded failure path. Validation returned that ranking for all five rejected outputs, avoiding repair prompts and retries. The checks preserve candidate coverage; clinician review determines whether the surfaced order is appropriate. Changes to catalogs, prompts, or candidate-set size can alter score distributions, so weights, formatting, threshold, validation, and fallback should be versioned together. Per-request scores, route, token counts, and stage timings support monitoring and shadow evaluation (Appendices~\ref{app:serving-interface}--\ref{app:prompts}). A shorter identifier-only output is a promising efficiency comparison, but would require measuring its ranking quality and validation behavior alongside its decoding cost.

\section{Conclusion}
\label{sec:conclusion}

\ogrank{} uses the same candidate scores to rank clinical orders and decide when to generate a joint ranking. This selective policy captures most of full generation's improvement in signed-order agreement while avoiding generation for a majority of requests. Its advantage over margin routing at matched call budgets shows that the allocation rule contributes to the result. Score reuse thus offers a simple way to improve the quality--latency tradeoff of a reranking component while retaining a complete fast ranking for every request.

\section*{Limitations}

This study evaluates a fixed checkpoint within one operational EHR workflow. Training/evaluation exclusion is established at case level; the patient-clustered intervals account for dependence within the test set, rather than establishing unseen-patient or cross-site performance. Target-containing lists isolate reranking from retrieval, and excluding the nine target replacements preserves the ranking gain. The signed order may not be the only clinically appropriate choice. Entropy measures score concentration, and the content-based difficulty annotations characterize a separate aspect of the candidate lists. Latency applies to the measured reranker and serving configuration; prospective workflow effects require evaluation with clinicians. The staff diagnostic and automated reward rubric are not clinical adjudication.

\section*{Ethical Considerations}

This retrospective offline evaluation used de-identified cases; research artifacts contained no protected health information (PHI), remained in a secured environment, and were protected by access controls, encryption at rest, and audit logging.
No re-identification was attempted, data use followed applicable institutional agreements and governance processes, and no model output affected care.
GPT-4o derived order intents without access to the signed-order targets, and GPT-5 supplied offline optimization rewards. Neither model supplied evaluation labels. GPT-5 did not participate in serving; the reported reranker timings exclude the upstream GPT-4o call.
The clinical data and adapted weights are not released.
Clinical order selection is consequential, and a plausible but incorrect ranking can encourage omission, under-specification, or automation bias, meaning over-reliance on system suggestions.
\ogrank{} only surfaces supplied candidates for clinician review and cannot create, authorize, or transmit an order.
Structural checks enforce format and candidate coverage, while clinical appropriateness still requires clinician review.
The reported endpoints measure workflow-label agreement, not patient benefit or clinical safety.

The signed-order target may encode local practice and documentation biases as well as catalog structure.
Each target site, patient population, and catalog therefore requires site-specific validation.
Any prospective use should include site-approved governance, performance auditing across patient groups, drift monitoring, logging of escalations and overrides, and workflow evaluation that preserves meaningful clinician control.

\bibliography{references}

\appendix
\raggedbottom

\setcounter{topnumber}{3}
\setcounter{dbltopnumber}{3}
\setcounter{totalnumber}{5}
\renewcommand{\topfraction}{0.92}
\renewcommand{\dbltopfraction}{0.92}
\renewcommand{\textfraction}{0.06}

\section{Evaluation Corpus and Protocol}
\label{app:data-profile}

\subsection{Study unit and data boundary}

The evaluation unit is one ordering case: de-identified encounter text and available patient context, a clinician-signed finalized order, which we call the signed-order target, and an encounter-scoped set of local-catalog candidates.
All 214 evaluation cases were excluded from the approximately 500 source encounters used to construct training prompts.
The primary test set contains 187 unique patients and enforces case-level exclusion from adaptation. Development and evaluation are patient-disjoint. This split definition does not establish patient disjointness between adaptation and evaluation; two additional same-workflow evaluation partitions are described in Appendix~\ref{app:holdouts}.
Outputs were evaluated offline and did not affect care.
All ranking analyses estimate performance conditional on candidate inclusion; performance of the upstream retrieval component is outside the quantity estimated here.

\begin{figure*}[t]
  \centering
  \includegraphics[width=0.90\textwidth]{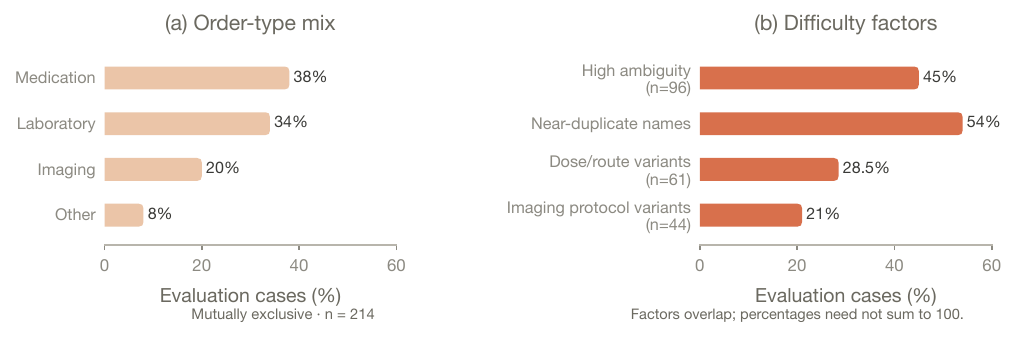}
  \caption{\textbf{Evaluation case-subset profile ($n=214$).}
  Left: mutually exclusive order types.
  Right: overlapping difficulty factors; percentages therefore need not sum to 100.}
  \label{fig:data-profile}
\end{figure*}

Figure~\ref{fig:data-profile} summarizes corpus composition.
The mutually exclusive order-type partition is 38\% medication, 34\% laboratory, 20\% imaging, and 8\% other.
The overlapping difficulty annotations identify 96 high-ambiguity cases (45\%), near-duplicate names in 54\% of cases, dose/route variants in 61 (28.5\%), and imaging protocol variants in 44 (21\%).
Consequently, neither the difficulty percentages nor the case-subset sample sizes should be summed. These content annotations are defined separately from the entropy gate, which uses the model scores.

\subsection{Candidate construction and target-inclusion control}

At evaluation time, GPT-4o converts encounter context to a free-text order intent without access to the finalized signed order. In the primary set, 62\% of intents name a concrete medication, test, or imaging examination; 38\% describe a broader clinical goal.
An upstream retrieval component then returns the 20 highest-ranked catalog-eligible candidates.
The signed-order target occurs naturally in approximately 96\% of these lists.
For the remaining nine lists (4.2\%), the final retrieved candidate is replaced by the target, keeping $N=20$.
This operation changes the candidate pool but supplies no explicit target label to the reranker.
It defines a controlled comparison of ranking policies on identical target-containing lists.
Removing these nine lists changes Recall@1 for fast, entropy, and full slow by at most .009. Entropy retains a net gain of 23 matches over fast on the remaining 205 lists. Because routing scores the supplied list, this sensitivity analysis removes the affected lists rather than reconstructing missing-target rankings.

\subsection{Metrics and inference modes}

There is one signed-order target per list.
If its returned rank is $r$, the per-case top-1 match indicator and nDCG@20 value are
\begin{align}
M(r) &= \mathbb{I}[r=1], \\
\mathrm{nDCG@20} &=
\begin{cases}
1/\log_2(r+1), & r\leq20,\\
0, & r>20.
\end{cases}
\end{align}
Corpus Recall@1 is the arithmetic mean of $M(r)$, and corpus nDCG@20 is the arithmetic mean of its per-case value.
Here, $\mathbb{I}$ is an indicator: it equals 1 when the condition holds and 0 otherwise.
Because the target-inclusion control ensures $r\leq20$, nDCG@20 measures where the signed-order target appears within the supplied list.
Neither metric gives credit to a clinically plausible alternative.

\emph{Fast} sorts the first-position scores in Equation~\ref{eq:fast-score} without decoding.
\emph{Two-speed} applies the frozen $T=0.90$ gate and accepts one validated complete ranking when invoked.
\emph{Full slow} invokes the same bounded joint-ranking generation for every list.
The full slow column is the reference that invokes this call for every list.

\subsection{Development-set threshold selection}
\label{app:threshold-selection}

We selected the routing threshold once on a 200-case development set containing no evaluation-set patients.
Before test evaluation, we prespecified a maximum development-set slow-path fraction of 45\% and selected the threshold that maximized development Recall@1 subject to that constraint.
We chose the cap as a study budget that requires a majority fast path while retaining substantial refinement coverage; it is not an estimated optimum, a measured service-level objective, or a recommended deployment budget.
Ties were resolved first by lower slow-path use and then by higher nDCG@20.
This procedure selected $T=0.90$, which was frozen and applied unchanged to every reported evaluation set.
The entropy policy's test-set slow-path fractions were observed consequences of the frozen threshold and were neither matched nor retuned.
On the primary test set, the margin gate was assigned the same invocation count; the uniform-allocation reference conditions on that count.

Across 1,000 patient-level bootstrap resamples, each formed by sampling patients with replacement, the median threshold estimate was 0.90, with a 95\% interval of $[0.88, 0.92]$.
The slow-path fractions associated with those resampled estimates ranged from 41\% to 47\%.
The 45\% constraint governed the single selection on the full development set; bootstrap values describe sampling sensitivity around that frozen decision and were not used to revise it.
This bootstrap analysis describes threshold selection rather than test performance. Separately selecting development thresholds under 40\% and 50\% call caps changes primary Recall@1 by at most .014 and preserves the router ordering. The primary $T=0.90$ policy remains unchanged.

\subsection{Configuration evidence roles and timing boundary}

Table~\ref{tab:model-artifacts} identifies each checkpoint and the evidence reported for it.

\begin{table*}[t]
\centering
\small
\renewcommand{\arraystretch}{1.12}
\setlength{\tabcolsep}{4pt}
\begin{tabular}{@{}>{\raggedright\arraybackslash}p{0.24\textwidth}>{\raggedright\arraybackslash}p{0.29\textwidth}>{\raggedright\arraybackslash}p{0.38\textwidth}@{}}
\toprule
\rowcolor{TableHeaderShade}
\textbf{Configuration} & \textbf{Base model or checkpoint} & \textbf{Evidence reported in this paper} \\
\midrule
\rowcolor{TableGroupShade}
\multicolumn{3}{@{}l}{\textit{Primary quantitative evidence}} \\

PubMedBERT bi-encoder
& PubMedBERT \cite{gu2021pubmedbert}
& Primary 214-list no-decoding context: Recall@1 .13; nDCG@20 .364. \\
MS-MARCO MiniLM-L12-v2
& MS-MARCO MiniLM-L12-v2 model card \cite{sentenceTransformers2026minilm}
& Fine-tuned on the same clinical training data as \ogrank{}; primary 214-list Recall@1 .18 and nDCG@20 .437. \\
Llama-3.2-3B-Instruct
& Llama-3.2-3B-Instruct model card \cite{meta2024llama32}
& Public-checkpoint fast result in Table~\ref{tab:training-ablation}; no routing claim. \\
\ogrank{} 3B
& LoRA-adapted Llama-3.2-3B-Instruct
& Primary test-set fixed-checkpoint fast, two-speed, and full slow analysis in Table~\ref{tab:main_results}. \\
\midrule
\rowcolor{TableGroupShade}
\multicolumn{3}{@{}l}{\textit{Targeted 40-case staff diagnostic only}} \\

Qwen2.5-3B-Instruct
& Qwen2.5 instruction model \cite{qwen2024technical}
& Diagnostic configuration in Figure~\ref{fig:internal-review}. \\
MedGemma-4B-IT
& MedGemma 4B instruction model \cite{sellergren2025medgemma}
& Diagnostic configuration in Figure~\ref{fig:internal-review}. \\
Qwen3-Reranker-4B
& Qwen3 reranking model \cite{zhang2025qwen3embedding}
& Diagnostic configuration in Figure~\ref{fig:internal-review}. \\
\bottomrule
\end{tabular}
\caption{\textbf{Configuration inventory and evidence scope.}
Rows are grouped by the evidence reported in this paper.
The two encoder metrics use the 214 constructed target-containing lists in the primary test set and signed-order metrics.
Diagnostic-only rows identify configurations shown in Figure~\ref{fig:internal-review}; they do not imply primary-set evaluation or a matched baseline.}
\label{tab:model-artifacts}
\end{table*}

\Needspace{7\baselineskip}
Let $G$ denote the displayed slow-path fraction, or share of lists receiving a complete-ranking LLM call. Complete-ranking accounting uses $G$ and the reported 4.687\,s decoded time per full slow list.
Table~\ref{tab:operating-curve-appendix} reports $G\times4.687$\,s as an approximate slow-path-weighted value.
This descriptive calculation excludes retrieval, candidate-wise scoring, entropy computation, parsing, validation, transport, queueing, and percentile latency; cross-model latency comparisons require a matched execution environment.

\subsection{Additional same-workflow evaluation sets}
\label{app:holdouts}

Additional Sets A ($n=250$) and B ($n=300$) apply the frozen policy to supplied lists from the same operational setting.
These sets test whether the ranking trend recurs within the same operational workflow.
Table~\ref{tab:holdouts} reports the paired fast and two-speed values for all three evaluation sets. At matched call budgets, margin adds 16 matches over fast on A and 18 on B, compared with entropy's 25 and 28 (Table~\ref{tab:main_results}). 

\begin{table}[t]
\centering
\small
\renewcommand{\arraystretch}{1.12}
\setlength{\tabcolsep}{3pt}
\begin{tabular}{@{}lrrrrr@{}}
\toprule
\rowcolor{TableHeaderShade}
\textbf{Set} & \textbf{$n$} & \textbf{Fast} & \shortstack{\textbf{Two-}\\\textbf{speed}} & \textbf{Net} & \textbf{Slow \%} \\
\midrule
Primary          & 214 & .449 & .561 & +24 & 45 \\
Additional Set A & 250 & .428 & .528 & +25 & 43 \\
Additional Set B & 300 & .417 & .510 & +28 & 46 \\
\bottomrule
\end{tabular}
\caption{\textbf{Recall@1 under three evaluation partitions.}
Fast and two-speed columns report Recall@1; Net is the additional top-1 matches and Slow the slow-path fraction.
The additional evaluations are reported as paired point estimates; the primary aggregate paired agreement table supports the inferential test in Appendix~\ref{app:outcome-ledger}.}
\label{tab:holdouts}
\end{table}

\section{Paired Outcomes and Error Taxonomy}
\label{app:outcome-ledger}

\subsection{Matched-budget routing and operating curve}

All routing policies in Table~\ref{tab:routing-appendix} reuse the same \ogrank{} checkpoint, candidate-wise scores, candidate lists, and bounded slow-path ranking procedure.
The frozen entropy threshold invokes generation on 96/214 primary lists; the margin gate receives exactly 96 calls.
The uniform-allocation row is the expected result when exactly that many calls are assigned independently of the scores, while the margin gate selects lists with the smallest gaps between the two highest fast scores.
Patient-clustered paired intervals compare entropy with both margin and fast below.

\begin{table}[t]
\centering
\small
\renewcommand{\arraystretch}{1.12}
\setlength{\tabcolsep}{3pt}
\begin{tabular}{@{}lrrr@{}}
\toprule
\rowcolor{TableHeaderShade}
\textbf{Policy} & \textbf{Slow \%} & \textbf{Recall@1} & \textbf{Net} \\
\midrule
Fast only               & 0   & .449 & 0 \\
Uniform allocation       & 44.9 & .520 & +15.25 \\
Margin gate             & 44.9 & .519 & +15 \\
\rowcolor{MethodBlue}
\textbf{Two-speed (entropy gate)} & 44.9 & .561 & +24 \\
Full slow               & 100 & .607 & +34 \\
\bottomrule
\end{tabular}
\caption{\textbf{Controlled fixed-checkpoint routing point estimates.}
Slow is slow-path fraction, and Net is the signed-order-target top-1 count difference from fast over 214 lists.
Uniform allocation is the expectation over all assignments rather than one sample; margin and entropy are observed.}
\label{tab:routing-appendix}
\end{table}

Let $N=214$ and $m=96$. For case $i$, let $F_i,S_i\in\{0,1\}$ indicate whether fast and full slow, respectively, rank the target first, and let $A$ be a uniformly selected subset of exactly $m$ cases. Define the resulting policy's Recall@1 as
\begin{equation}
\mu(A)=\frac{1}{N}\sum_{i=1}^{N}\left(\mathbb{I}[i\in A]S_i+\mathbb{I}[i\notin A]F_i\right).
\end{equation}
Let $\mathbf{F}=(F_i)_{i=1}^{N}$ and $\mathbf{S}=(S_i)_{i=1}^{N}$ denote the fixed paired outcome vectors. Because $\Pr(i\in A)=m/N$,
\begin{align}
\mathbb{E}[\mu(A)\mid \mathbf{F},\mathbf{S}]
&=\frac{1}{N}\sum_i F_i
+\frac{m}{N^2}\sum_i(S_i-F_i) \notag\\
&=\frac{96}{214}+\frac{m(130-96)}{214^2}
\approx .52.
\end{align}
At $m=96$, the expected number of matches is $96+96(130-96)/214=111.25$ (rounded). This expectation avoids random variation from one sampled allocation.
It is a matched-budget reference, not an uncertainty interval or an inferential comparison with entropy.

\begin{table*}[t]
\centering
\small
\renewcommand{\arraystretch}{1.12}
\setlength{\tabcolsep}{5pt}
\begin{tabular}{@{}lrrrrr@{}}
\toprule
\rowcolor{TableHeaderShade}
\textbf{Entropy operating point} & \textbf{Slow path \%} & \textbf{Recall@1} & \textbf{nDCG@20} & \textbf{Net matches} & \shortstack{\textbf{Fraction-weighted}\\\textbf{decoded time}} \\
\midrule
Fast only        & 0   & .449 & .625 & 0   & 0 \\
20\% slow-path point & 20  & .514 & .663 & +14 & $\approx 0.94$\,s \\
\rowcolor{MethodBlue}
\textbf{Frozen ($T=0.90$)} & 45  & .561 & .699 & +24 & $\approx 2.11$\,s \\
65\% slow-path point & 65  & .579 & .719 & +28 & $\approx 3.05$\,s \\
Full slow       & 100 & .607 & .736 & +34 & 4.687\,s \\
\bottomrule
\end{tabular}
\caption{\textbf{Observed ranking quality versus slow-path use.}
Fraction-weighted decoded time multiplies each displayed rounded slow-path fraction by the reported 4.687\,s decoded time per full slow list; intermediate values are approximate complete-ranking estimates, not directly timed end-to-end request latency.}
\label{tab:operating-curve-appendix}
\end{table*}

\subsection{Patient-clustered paired uncertainty}

The primary evaluation has 214 cases from 187 patients. A 1,000-resample paired bootstrap groups cases by patient, so the policy comparisons retain within-patient dependence. Entropy exceeds margin by nine matches, with a 95\% interval of $[+2,+16]$, and exceeds fast by 24, with interval $[+15,+32]$. These intervals use the same fixed checkpoint, candidate lists, and routing outputs. They address dependence within the evaluation set; the training/evaluation split remains defined at case level.

\subsection{Paired fast-to-two-speed transitions}

Aggregate Recall@1 hides whether escalation creates or removes agreement with the signed-order target.
Figure~\ref{fig:outcome_analysis}(a) therefore reports the complete aggregate paired agreement table for the 214 lists.
Fast inference matches the signed-order target on 96 cases and two-speed inference on 120, yielding the reported rounded values of 0.45 and 0.56.
The net change is $30-6=24$ additional matches.

Only 36 cases change top-rank agreement between the two modes, called off-diagonal pairs.
The other 178 cases retain their top-rank agreement status, although their lower ranks may change.
The 5:1 gain-to-loss ratio is a descriptive paired count.
The original exact two-sided McNemar test on these 36 changed-status cases gives $p=6.96\times10^{-5}$ under case independence. The patient-clustered interval above is the uncertainty estimate used for the main result.

\subsection{Residual-error coding}

After obtaining the results, a review assigned one primary operational category to each of the 94 residual top-1 misses.

\Needspace{4\baselineskip}
This qualitative diagnosis used no formal adjudication or agreement protocol.
Counts were 28 multiple-plausible-alternative cases (29.8\%), 22 dose/route-specificity cases (23.4\%), 15 imaging-protocol/modality cases (16.0\%), 14 retrieval/catalog-label mismatch cases (14.9\%), 9 insufficient-context cases (9.6\%), and 6 slow-output formatting/fallback cases (6.4\%); percentages are rounded independently.

Multiple plausible alternatives motivate evaluation labels that can credit more than one answer rather than an automatic model-error conclusion.
Dose/route and imaging categories identify specificity issues in ranking or catalog representation, while insufficient context points to input assembly.

\Needspace{6\baselineskip}
The retrieval/catalog-label category combines candidate-construction and alignment with the signed catalog label, so it is not a measure of retrieval recall.
The formatting/fallback category counts residual top-1 misses, not request-level fallback frequency.

\begin{figure*}[t]
  \centering
  \includegraphics[width=0.96\textwidth]{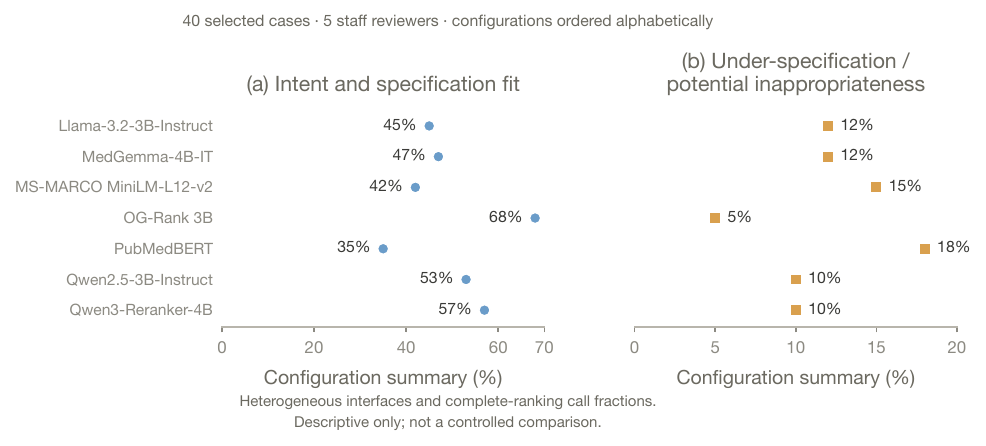}
  \caption{\textbf{Targeted staff diagnostic on 40 selected high-ambiguity cases.}
  Configurations use heterogeneous ranking interfaces and complete-ranking call fractions.
  Marks are rounded configuration-level summaries from five staff reviewers: three medical scribes and two data scientists.
  This deliberately selected diagnostic is separate from the primary test-set metrics in Table~\ref{tab:main_results}; it does not estimate model-family superiority, clinician preference, or clinical safety.}
  \label{fig:internal-review}
\end{figure*}

\section{Targeted Staff Diagnostic}
\label{app:internal-review}

Forty cases were deliberately selected from the 96-case high-ambiguity subset to form a descriptive sample.
Three medical scribes and two data scientists reviewed each top-ranked candidate against the case context.

\Needspace{4\baselineskip}
Encoder configurations used their standard ranking interfaces; decoder configurations used configuration-specific reranking and complete-ranking policies.
The review uses two staff-review labels: \emph{intent and specification fit} and \emph{under-specification or potential inappropriateness}.
These are descriptive error-analysis labels, not clinician adjudication, clinician preference, or clinical safety endpoints.

\Needspace{3\baselineskip}
Figure~\ref{fig:internal-review} presents rounded panel summaries from the targeted review.
We report the configuration summaries for qualitative error inspection without ranking configurations or treating differences as estimated effects.
Interfaces and complete-ranking call fractions differ across configurations, so the values do not support a model-family ranking or isolate an effect of the base model; encoder metrics on the primary test set are reported in Table~\ref{tab:main_results}.
We treat these percentages only as panel-level descriptors and do not convert them to case counts or pooled-reviewer estimates.
Because case selection was targeted and reviewer agreement was not measured, these values support qualitative diagnosis only.
The paired 214-case comparison provides the formal statistical analysis.

\section{Training Configuration and Optimization Details}
\label{app:training-details}

\subsection{Training-instance construction}

For every clinician-signed training target $c^\star$, the training construction uses that target to identify supporting encounter text, derive a compact query, and assemble an encounter-scoped top-20 pool.
When $c^\star$ is absent, it replaces the final retrieved candidate.
During training, the signed-order target is used to construct candidate pools and labels.
During evaluation, the finalized order is withheld from intent derivation and is not marked as the target in reranker prompts or routing; it is used for target-inclusion control and offline metrics.
The reported results therefore isolate reranking on target-containing candidate sets and do not evaluate deployment-time intent derivation.

Each pool is expanded into 20 candidate-level prompts.
Prompt $u_j$ highlights candidate $c_j$, lists the other 19 candidates as compact alternatives, and receives label $y_j=\mathbb{I}[c_j=c^\star]$.
Thus, each target-specific pool contributes one positive and 19 negative prompts.
Each of roughly 500 source encounters contributes a 20-candidate pool, yielding approximately 10,000 candidate-level prompts.
The requested completion begins with a lowercase \texttt{yes} or \texttt{no} decision and continues with a short rationale.

\subsection{Optimization configuration}

We adapt Llama-3.2-3B-Instruct with LoRA for five training passes, or epochs.
Table~\ref{tab:training-config} summarizes the optimization configuration.
Table~\ref{tab:training-ablation} separates the stagewise fast-path configurations from the serving-mode comparison.
LoRA targets attention and multilayer perceptron (MLP) blocks.

\begin{table*}[t]
\centering
\small
\renewcommand{\arraystretch}{1.12}
\setlength{\tabcolsep}{5pt}
\begin{tabular}{@{}>{\raggedright\arraybackslash}p{0.23\textwidth}>{\raggedright\arraybackslash}p{0.69\textwidth}@{}}
\toprule
\rowcolor{TableHeaderShade}
\textbf{Field} & \textbf{Configuration} \\
\midrule
Base model and duration & Llama-3.2-3B-Instruct; 5 epochs \\
LoRA & Attention and MLP blocks; adapter rank 16; scaling factor $\alpha=32$; dropout 0.1 \\
Optimizer & AdamW; learning rate $10^{-6}$; weight decay 0 \\
Batching & Micro-batch 1; gradient accumulation 4 \\
Gradient clipping & Threshold 1.0 \\
Sequence handling & 2,048-token prompt cap; left padding to align the first generated position \\
Reference regularization & Reference checkpoint frozen at each epoch boundary; tokenwise Kullback-Leibler (KL) divergence weight $\beta=0.02$ \\
Difficulty-allocation cadence & 2 cycles per epoch; prompts reassigned at epoch boundaries \\
\bottomrule
\end{tabular}
\caption{\textbf{Training configuration for the evaluated checkpoint.}}
\label{tab:training-config}
\end{table*}

\begin{table*}[t]
\centering
\small
\renewcommand{\arraystretch}{1.12}
\setlength{\tabcolsep}{4pt}
\begin{tabular}{@{}lccc rr@{}}
\toprule
\rowcolor{TableHeaderShade}
\textbf{Configuration} & \shortstack{\textbf{Pointwise}\\\textbf{adaptation}} & \shortstack{\textbf{Rubric-guided}\\\textbf{update}} & \shortstack{\textbf{Inference}\\\textbf{mode}} & \textbf{Recall@1} & \textbf{nDCG@20} \\
\midrule
Llama-3.2-3B-Instruct (public) & No  & No  & Fast      & .25 & .496 \\
Supervised pointwise    & Yes & No  & Fast      & .40 & .590 \\
Rubric-guided pointwise & Yes & Yes & Fast      & .449 & .625 \\
\midrule
\rowcolor{MethodBlue}
\textbf{Two-speed policy} & Yes & Yes & Two-speed & .561 & .699 \\
Full slow               & Yes & Yes & Full slow & .607 & .736 \\
\bottomrule
\end{tabular}
\caption{\textbf{Stagewise checkpoints and fixed-checkpoint serving modes.}
Rows 1--3 trace successive fast-mode checkpoints and are not a factorial training ablation.
The third row is the shared fast anchor for rows 3--5, which hold the final adapted checkpoint fixed and compare serving modes.}
\label{tab:training-ablation}
\end{table*}

Training uses a Group Relative Policy Optimization (GRPO)-style objective with tokenwise KL regularization to a reference checkpoint frozen at the start of each epoch \cite{liu2024deepseekmath}.
Only generated tokens contribute to the policy update.

\subsection{Automated reward rubric}

For completion $t$, the offline training judge returns true/false checks in five dimensions: decision agreement, clinical relevance, specificity, safety, and format.

\Needspace{6\baselineskip}
The training implementation maps these checks to bounded dimension scores $S_d(t)\in[-3,3]$, where $d$ is one of the five named rubric dimensions; designated checks that force failure of a dimension assign it $-3$.
\begin{align}
R(t)={}&3.0S_{\mathrm{decision}}(t)
+1.5S_{\mathrm{clinical}}(t)\notag\\
&+1.5S_{\mathrm{specificity}}(t)
+2.0S_{\mathrm{safety}}(t)\notag\\
&+1.5S_{\mathrm{format}}(t).
\label{eq:app-reward}
\end{align}
GPT-5 supplies offline rubric rewards during optimization only \cite{openai2025gpt5}.
It does not define evaluation labels or participate in serving, and the rubric dimension named \emph{safety} is not a clinical safety assessment.

\subsection{Ambiguity- and judged-trend-guided curriculum}

For candidate-wise log-odds $z_j$, the training-allocation ambiguity score is
\begin{equation}
q(u_j)=4\,\sigma(z_j)\bigl(1-\sigma(z_j)\bigr).
\label{eq:pointwise-ambiguity}
\end{equation}
Here, $\sigma(a)=(1+\exp(-a))^{-1}$ is the logistic sigmoid. The score is recomputed at epoch boundaries and combined with the prompt's mean logged rubric score from the preceding epoch.
This candidate-level score is largest when the yes/no probability is near .5 and is distinct from the list-normalized entropy $U$ used for serving.
Fixed rules assign a prompt to hard when ambiguity is high or its prior score is low, to medium when either signal is intermediate, and to easy otherwise. Ambiguity cutoffs were derived from training-split quantiles, and reward thresholds were fixed on the $[-3,3]$ rubric scale.

Each bucket uses a fixed number of rollouts, meaning sampled training completions, plus fixed sampling and rationale-length settings (Table~\ref{tab:bucket-trace}).
The larger hard-bucket settings allocate more sampled training completions to ambiguous or persistently low-reward prompts.
Every sampled completion and its rubric scores were logged in JSON Lines (JSONL) format, enabling the next epoch's reward statistic and the aggregate shares shown here.

\begin{figure}[H]
  \centering
  \includegraphics[width=0.96\columnwidth]{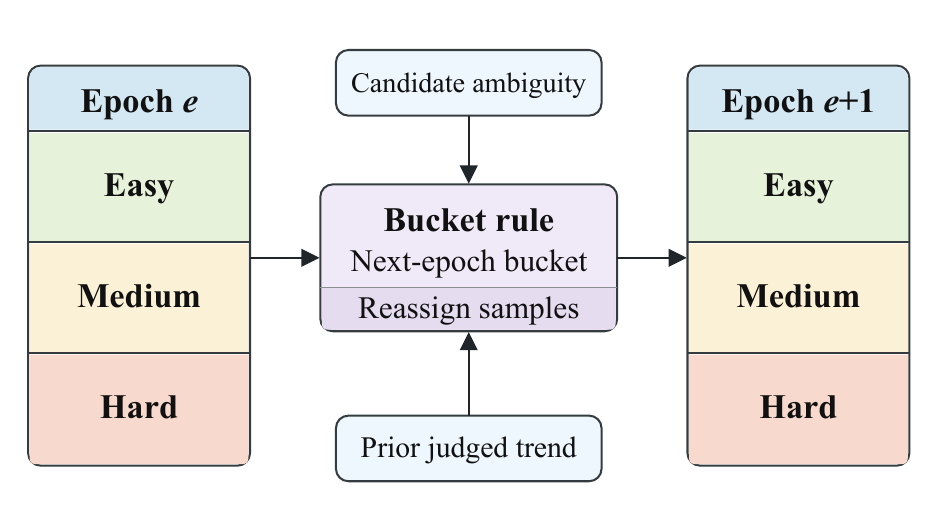}
  \caption{\textbf{Epoch-boundary prompt reassignment.}
  Candidate-level ambiguity and the preceding epoch's rubric trend reassign prompts among fixed easy, medium, and hard compute buckets.}
  \label{fig:rebucketing}
\end{figure}

\begin{figure*}[t]
  \centering
  \includegraphics[width=0.94\textwidth]{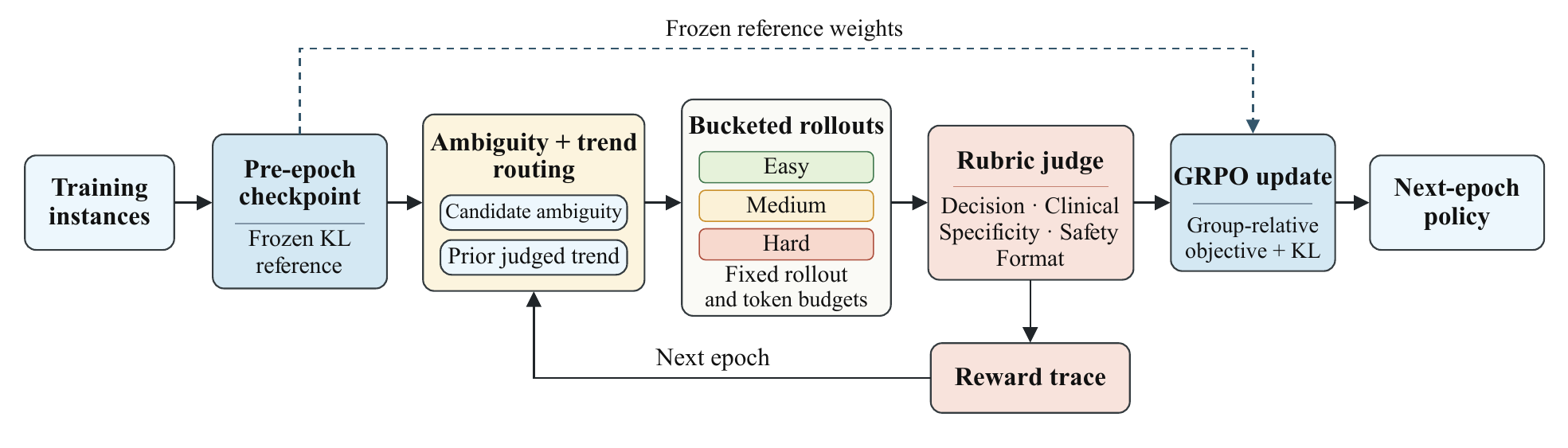}
  \caption{\textbf{Training loop used for the evaluated checkpoint.}
  Candidate-level decision-plus-rationale completions receive a fixed five-part rubric reward, group-relative updates, and reassignment among fixed compute buckets at epoch boundaries.
  The complete-ranking interface is used only at serving time.}
  \label{fig:training-loop}
\end{figure*}

Figure~\ref{fig:rebucketing} isolates the epoch-boundary reassignment rule, while Figure~\ref{fig:training-loop} summarizes the complete optimization loop.

\begin{table*}[t]
\centering
\small
\renewcommand{\arraystretch}{1.12}
\setlength{\tabcolsep}{4pt}
\begin{tabular}{@{}lrrrrrrrrr@{}}
\toprule
& \multicolumn{4}{c}{\textbf{Fixed bucket configuration}}
& \multicolumn{5}{c}{\textbf{Observed prompt share (\%)}} \\
\cmidrule(lr){2-5}\cmidrule(lr){6-10}
\rowcolor{TableHeaderShade}
\textbf{Bucket} & \textbf{Samples} & \textbf{Temperature} & \textbf{Top-$p$} & \textbf{Max. tokens}
& \textbf{Epoch 0} & \textbf{Epoch 1} & \textbf{Epoch 2} & \textbf{Epoch 3} & \textbf{Epoch 4} \\
\midrule
Easy   & 2 & .2 & .80 & 100 & 34.0 & 39.5 & 43.0 & 46.0 & 49.0 \\
Medium & 4 & .4 & .92 & 200 & 56.0 & 47.0 & 44.0 & 41.0 & 38.0 \\
Hard   & 6 & .7 & .97 & 300 & 10.0 & 13.5 & 13.0 & 13.0 & 13.0 \\
\bottomrule
\end{tabular}
\caption{\textbf{Curriculum completion-sampling budgets and epoch-level bucket composition.}
Samples are completions generated per prompt; temperature controls sampling randomness, and Top-$p$ is the cumulative-probability cutoff for token sampling. Prompt shares show how training allocation changes across epochs.}
\label{tab:bucket-trace}
\end{table*}

\subsection{Pointwise optimization and complete-ranking reuse}

Optimization is candidate-level: the model generates a binary decision followed by a rationale for one highlighted candidate.
It does not directly optimize the inference-time JSON permutation.
The slow path instead applies a separate prompt containing all supplied candidates to the same adapted weights.
Its ability to produce a complete candidate ordering therefore comes from reusing candidate-level trained weights under that prompt, not from direct supervision on complete orderings.

\section{Detailed Slice Metrics}
\label{app:slice-metrics}

The two slices provide complementary views of specification-sensitive ranking.
High ambiguity identifies 96 lists with multiple plausible alternatives, close dose/route or protocol variants, or other small specification differences.
Medication dose/route identifies 61 lists in which the candidate set itself contains dose or route distinctions.
Because a list can satisfy both definitions, the slices overlap and do not partition the full evaluation.
Table~\ref{tab:slices-complete} reports Recall@1 and nDCG@20 for each slice and for all 214 lists.

Within each subset, Recall@1 asks whether the signed-order target reaches the first position, while nDCG@20 also credits upward movement that stops below rank one.
Both remain single-target measures of historical workflow agreement.
Another clinically plausible candidate receives no credit, and slice membership does not alter the reference label.
The slice results therefore localize ranking behavior under the same fixed-list, target-inclusion design as the primary test set; they are not measurements of clinical appropriateness or safety.

Relative to \ogrank{} fast, two-speed improves high-ambiguity Recall@1 by .16 and nDCG@20 by .100.
For medication dose/route cases, Recall@1 rises from .43 to .57 and nDCG@20 increases by .098.
These within-slice differences describe the final rankings on fixed subsets; they do not establish that the gate selected exactly the cases that benefited from a slow-path ranking call.
They also should not be added across slices because the subsets overlap.

\begin{table}[H]
\centering
\small
\renewcommand{\arraystretch}{1.12}
\setlength{\tabcolsep}{3pt}
\begin{tabular}{@{}llrrr@{}}
\toprule
\rowcolor{TableHeaderShade}
\textbf{Mode} & \textbf{Metric} & \textbf{All} & \shortstack{\textbf{High}\\\textbf{ambiguity}} & \shortstack{\textbf{Dose/}\\\textbf{route}} \\
\rowcolor{TableHeaderShade}
& & $(n=214)$ & $(n=96)$ & $(n=61)$ \\
\midrule
Fast      & Recall@1 & .45  & .36  & .43  \\
          & nDCG@20 & .625 & .571 & .608 \\
\addlinespace[2pt]
\textbf{Two-speed} & Recall@1 & .56  & .52  & .57  \\
          & nDCG@20 & .699 & .671 & .706 \\
\bottomrule
\end{tabular}
\caption{\textbf{Paired \ogrank{} metrics for the overlapping descriptive slices.}
Recall@1 is rounded to two decimals and nDCG@20 to three; values are conditional on target inclusion, and no subgroup significance claim is made.}
\label{tab:slices-complete}
\end{table}

\section{Serving Interface}
\label{app:serving-interface}

\subsection{Request scope and prompt contracts}

Upstream retrieval limits the candidate set to catalog-eligible entries.
The reranker compares clinical intent, specification, and supplied patient constraints within that set; operational eligibility, cost, and availability are outside its scoring objective.
The service ranks a fixed candidate list derived from context and optional patient signals; the exact text representation of the local catalog is deployment-specific.
Table~\ref{tab:prompt-contracts} summarizes the functional prompt contracts, and Appendix~\ref{app:prompts} provides representative templates.

\begin{figure}[H]
  \centering
  \includegraphics[width=0.96\columnwidth]{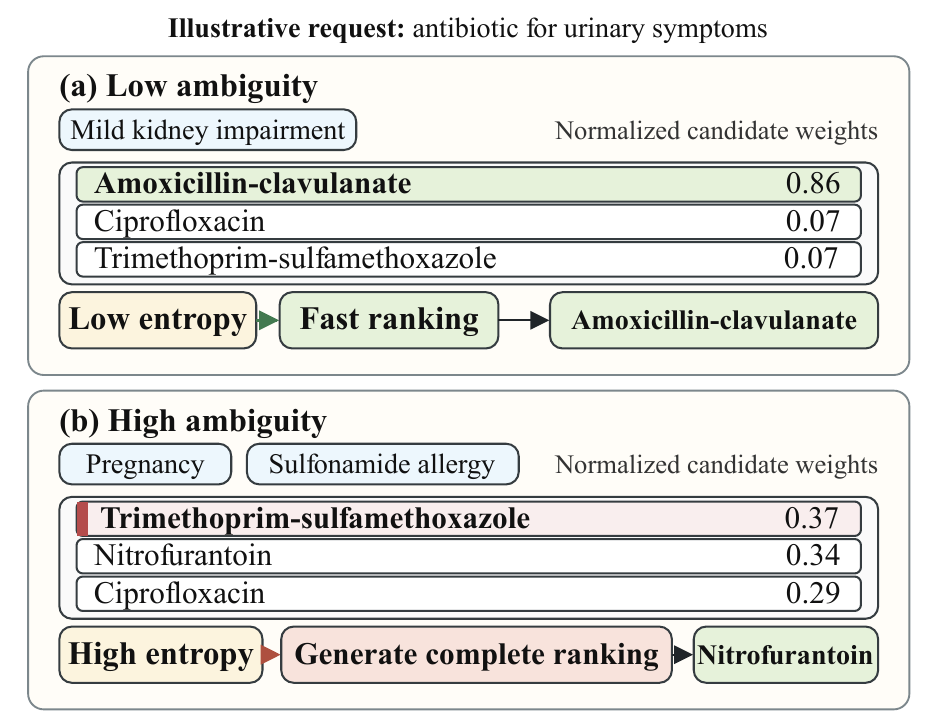}
  \caption{\textbf{Synthetic two-speed interface illustration.}
  Illustrative normalized candidate weights show a concentrated list returning the fast order and a diffuse list permitting one bounded slow-path call.
  Values are not correctness probabilities, empirical results, or clinical recommendations.}
\label{fig:two-speed-example}
\end{figure}

Figure~\ref{fig:two-speed-example} illustrates how diffuse first-step scores can route a list to one bounded slow-path call.

\begin{table}[H]
\centering
\small
\renewcommand{\arraystretch}{1.07}
\setlength{\tabcolsep}{2pt}
\begin{tabular}{@{}>{\raggedright\arraybackslash}p{0.20\columnwidth}>{\raggedright\arraybackslash}p{0.29\columnwidth}>{\raggedright\arraybackslash}p{0.43\columnwidth}@{}}
\toprule
\rowcolor{TableHeaderShade}
\textbf{Interface} & \textbf{Inputs} & \textbf{Required output and use} \\
\midrule
\textbf{Candidate-wise fast}
& Context, optional patient signals, one highlighted candidate, compact alternatives
& Serving reads pooled first-position \texttt{yes}/\texttt{no} unnormalized scores, or logits, to produce $z_j$ and generates no text. Candidate-level training uses the separate decision-plus-rationale contract below. \\
\midrule
\textbf{Slow-path joint ranking}
& Same context and patient fields; all candidates in the fixed list
& One JSON object requesting a ranking of the supplied candidates and a brief rationale. The ranking is considered only if it parses and exactly covers the supplied candidates. \\
\midrule
\textbf{Training judge}
& Candidate-level prompt, binary reference label, and sampled completion
& True/false rubric checks aggregated into the five reward dimensions in Equation~\ref{eq:app-reward}; never used as a test label or serving-time adjudicator. \\
\bottomrule
\end{tabular}
\caption{\textbf{Functional prompt contracts.}
The rationale field is auxiliary generated text and does not affect ranking acceptance.}
\label{tab:prompt-contracts}
\end{table}

Fast scoring uses the fixed, disjoint yes/no token sets in Equation~\ref{eq:fast-score} and one candidate-wise evaluation per candidate; the 20 evaluations may be batched.

\begin{figure}[H]
  \centering
  \includegraphics[width=0.98\columnwidth]{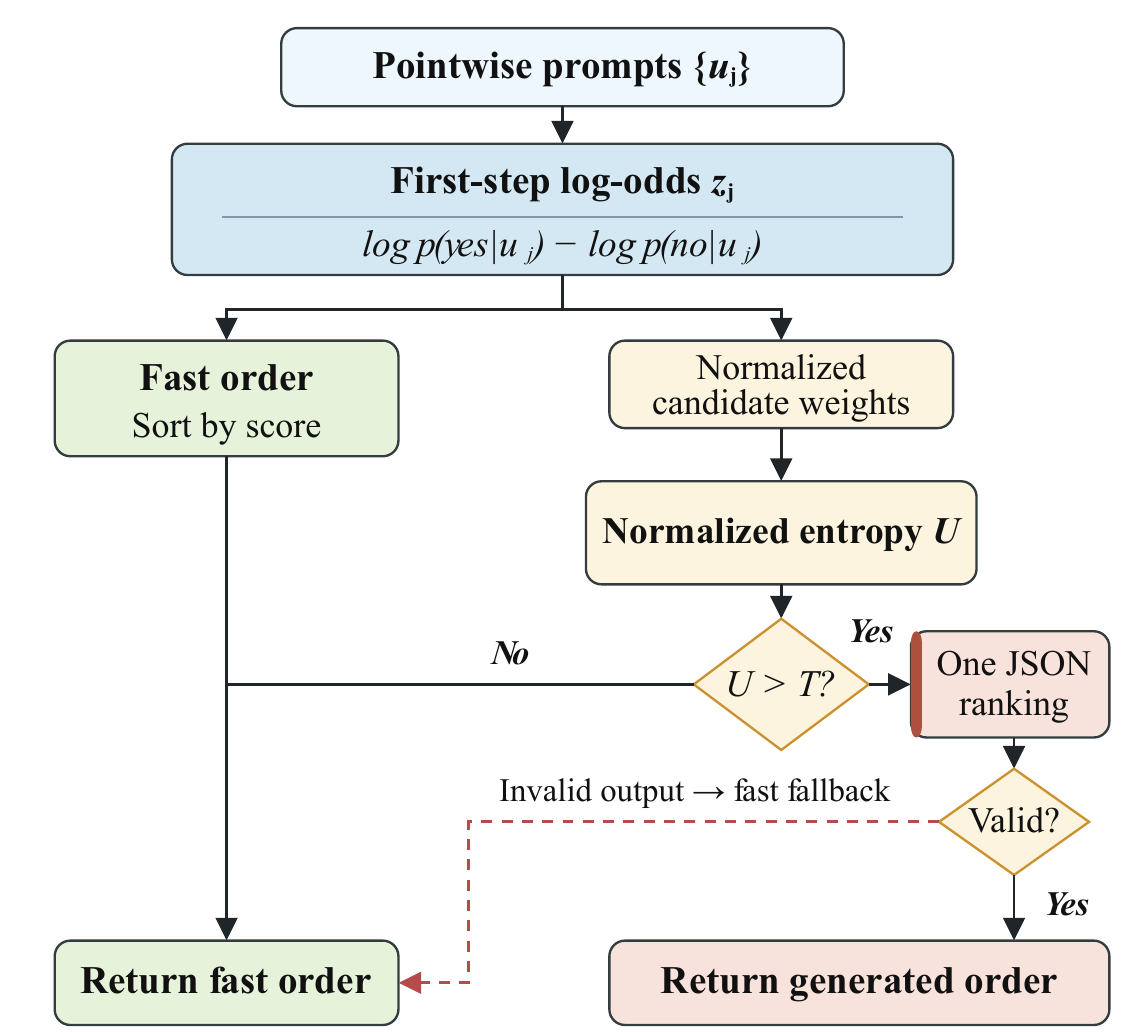}
  \caption{\textbf{Notation-focused inference companion.}
  Yes/no abbreviates the pooled token evidence in Equation~\ref{eq:fast-score}.
  The scores produce the fast order and normalized entropy; a low-entropy or invalid generated-ranking path returns the fast order, while the other path returns one valid JSON ranking.}
\label{fig:inference-notation}
\end{figure}

Figure~\ref{fig:inference-notation} summarizes the return paths using the notation of Equations~\ref{eq:fast-score}--\ref{eq:serving-policy}.

\subsection{Slow-path bound and validation}

If $U>0.90$, the service sends the context and all supplied candidates to the same checkpoint in one complete-ranking request with greedy decoding and a 512-token output cap.
There is no repair prompt, retry loop, or second complete-ranking attempt.
The requested object contains a ranking field and a brief rationale field.
Generated rankings were accepted only when the response parsed as the requested JSON object.
The returned ranking had to cover and resolve every supplied candidate.
Any parse, coverage, or candidate-resolution failure rejected the generated ranking.

\Needspace{7\baselineskip}
The rationale is auxiliary text and does not affect ranking acceptance.
Appendix~\ref{app:slow-prompt} shows the representative JSON structure. Exact candidate rendering, normalization, and tokenizer-specific decision-token variants belong to the versioned serving configuration.

\Needspace{14\baselineskip}
\subsection{Bounded control flow}

The service computes and saves the complete fast ranking before routing.
If $U\leq0.90$, it returns that ranking.
Otherwise, it makes one bounded slow-path call and returns the generated ordering only if validation succeeds; every rejected response returns the saved fast ranking.
This control flow bounds complete-ranking attempts, while clinical appropriateness remains subject to clinician review.

\FloatBarrier

\subsection{Measured configuration and validation}

Table~\ref{tab:serving-config} collects the serving settings underlying the latency comparison. The mean input length is 1,142 tokens; accepted slow outputs average 187 tokens. These summaries describe the measured requests, rather than throughput under concurrent load.

\begin{table}[tb]
\centering
\small
\renewcommand{\arraystretch}{1.12}
\setlength{\tabcolsep}{3pt}
\begin{tabular}{@{}>{\raggedright\arraybackslash}p{.36\columnwidth}>{\raggedright\arraybackslash}p{.58\columnwidth}@{}}
\toprule
\rowcolor{TableHeaderShade}
\textbf{Setting} & \textbf{Measured configuration} \\
\midrule
Accelerator & One NVIDIA A100 80\,GB \\
Software & CUDA 12.2; vLLM 0.6.1 \\
Request concurrency & 1 \\
Fast stage & 20 candidate prompts batched \\
Slow decoding & Greedy; 512 output tokens maximum \\
Attempts & One; no repair or retry \\
Mean input length & 1,142 tokens \\
Mean accepted output & 187 tokens \\
\bottomrule
\end{tabular}
\caption{\textbf{Serving configuration for reranker timing.} The interval excludes upstream intent derivation and catalog retrieval.}
\label{tab:serving-config}
\end{table}

\begin{table}[tb]
\centering
\small
\renewcommand{\arraystretch}{1.12}
\setlength{\tabcolsep}{4pt}
\begin{tabular}{@{}lrr@{}}
\toprule
\rowcolor{TableHeaderShade}
\textbf{Primary slow-call outcome} & \textbf{Requests} & \textbf{Share} \\
\midrule
Valid complete ranking & 91 & 94.8\% \\
Parsing failure & 3 & 3.1\% \\
Incomplete ranking & 2 & 2.1\% \\
\midrule
\textbf{All attempts} & \textbf{96} & \textbf{100\%} \\
\bottomrule
\end{tabular}
\caption{\textbf{Validation and fallback over 96 slow calls.} All five rejected outputs return the saved fast ranking. Additional-set validation success is 93--96\%.}
\label{tab:validation}
\end{table}

The five rejections in Table~\ref{tab:validation} are request-level validation outcomes. The six formatting/fallback labels in the residual-error taxonomy instead classify misses during qualitative review and do not estimate this failure rate.

\subsection{Reranking latency}

Table~\ref{tab:e2e-latency-appendix} measures complete reranking latency for the 214 requests in one execution environment.
Unlike slow-path-weighted decoded time, it includes the always-run candidate-wise stage and the remaining reranker control flow.
It begins with the supplied reranking input rather than upstream retrieval, so it is an end-to-end measurement of the reranker, not the broader clinical application.

\begin{table}[H]
\centering
\small
\renewcommand{\arraystretch}{1.12}
\setlength{\tabcolsep}{3pt}
\begin{tabular}{@{}lrrrrr@{}}
\toprule
\rowcolor{TableHeaderShade}
\textbf{Mode} & \shortstack{\textbf{Slow}\\\textbf{path \%}} & \textbf{Recall@1} & \textbf{p50} & \textbf{p95} & \textbf{Mean} \\
\midrule
Fast only  & 0   & .449 & 0.4\,s & 0.8\,s & 0.5\,s \\
\rowcolor{MethodBlue}
\textbf{Two-speed} & 45  & .561 & 0.6\,s & 5.8\,s & 2.7\,s \\
Full slow & 100 & .607 & 4.9\,s & 6.2\,s & 5.0\,s \\
\bottomrule
\end{tabular}
\caption{\textbf{Measured reranking latency.}
Slow path is the fraction of lists receiving complete-ranking calls; p50 is the median and p95 the 95th percentile. Values characterize one execution environment and are not cross-hardware benchmarks.}
\label{tab:e2e-latency-appendix}
\end{table}

\subsection{Measured and required runtime records}
\label{app:runtime-records}

Study measurements comprise signed-order-target ranks for Recall@1, nDCG@20, case subsets, and paired transitions; aggregate slow-path fractions; complete-ranking values derived from the 4.687\,s full slow reference; and end-to-end p50, p95, and mean for all three modes in one environment.
The ranks measure workflow-label agreement, not correctness; one threshold need not yield matched slow-path fractions across datasets; derived complete-ranking values exclude other request stages; and the end-to-end values are not cross-hardware benchmarks.

Aggregate validation and fallback counts are reported in Table~\ref{tab:validation}. A per-request runtime record should retain those outcomes and reason codes; model, prompt, schema, catalog, and threshold versions; candidate identifiers and scores; entropy; returned path; token count; stop reason; and stage timings.
Clinician selection or override should remain separate from model scores in prospective evaluation.

\section{Reproducibility Scope}
\label{app:reproducibility}

\paragraph{Policy-level reimplementation.}
Equations~\ref{eq:fast-score}--\ref{eq:serving-policy}, the frozen threshold-selection rule, candidate construction, metrics, and bounded control flow specify how to reimplement routing and evaluation for a compatible checkpoint.

\paragraph{Exact model-level reproduction.}
Exact reproduction additionally requires the governed clinical data and adapted weights; retrieval and intent derivation; the exact model-specific formatting of messages and tokens; decision-token sets; candidate matching, validation, and decoding configuration; optimization and judge configuration; and software, seeds, hardware, and file hashes.
Governed clinical data, adapted weights, and the complete version record are not released; the paper therefore supports policy-level reimplementation, not exact model-level reproduction or cross-hardware latency replication.

\section{Prompt Interface Specifications}
\label{app:prompts}

This appendix presents representative templates for four functional interfaces: fast scoring, a bounded slow-path call, candidate-level adaptation, and reward judging.
Exact model-specific message formatting, tokenizer rendering, and local field assembly must be versioned for reproduction, as detailed in Appendix~\ref{app:reproducibility}; Appendix~\ref{app:serving-interface} defines serving control flow.
Serving does not decode the fast-path decision token; the template fixes the first-position token choices whose logits are scored.
Table~\ref{tab:prompt-contracts} and Appendix~\ref{app:runtime-records} distinguish serving inputs, adaptation-only supervision, and the metadata needed to reproduce an operational decision.

\subsection{Fast-path candidate-wise scoring template}
\label{app:fast-prompt}

\begin{promptbox}{Fast-path prompt: inputs and task}
You are a clinical order relevance judge.
Inputs:
- CONTEXT: encounter text expressing clinician ordering intent.
- PATIENT: optional age, sex, pregnancy, allergies, medications, history, organ function, devices, and anticoagulation.
- CANDIDATE_TO_EVALUATE: one candidate order to judge.
- OTHER_CANDIDATES: all remaining supplied candidates.
Task:
Decide whether CANDIDATE_TO_EVALUATE is the best initial order for CONTEXT among all supplied candidates.
Check intent, safety, and required specificity: imaging type/body region/acquisition protocol; diagnostic target for laboratory tests; dose/administration route/frequency for medications.
Use PATIENT only when clinically relevant. Do not invent patient facts.
\end{promptbox}

\begin{promptbox}{Fast-path prompt: decision and output rules}
Best-of-pool rule:
- Answer yes only if the candidate is clearly best.
- Answer no for a tie without a uniquely best option.
- Answer no if unsafe, mismatched, or under-specified relative to a supplied alternative.
Ignore logistics, utilization, cost, availability, and non-clinical ordering history.
Output format:
- Output exactly one lowercase token: yes or no.
\end{promptbox}

\subsection{Bounded joint-ranking generation template}
\label{app:slow-prompt}

\begin{promptbox}{Slow-path system prompt: one JSON ordering of all candidates}
You are a clinical order relevance judge.
Given CONTEXT, optional PATIENT, and CANDIDATES:
1. Rank every candidate from best to worst.
2. Include a brief rationale.
Penalize intent mismatch, unsafe medication, and under-specification. Break ties by safety and specificity.
Output strictly one JSON object with no prose before or after it:
{
  "ranking": [
    {"text": "<candidate string>", "rank": 1}
  ],
  "rationale": "<brief explanation>"
}
\end{promptbox}

\begin{promptbox}{Serving-side response checks: not prompt text}
Accept the generated ranking only if:
- the response is one parseable JSON object;
- every ranking entry resolves to a supplied candidate; and
- the ranking covers the supplied candidate set.
Otherwise, return the precomputed fast ranking. Do not repair or retry.
\end{promptbox}

\subsection{Candidate-level training template}
\label{app:train-prompt}

\begin{promptbox}{Training prompt: inputs and task}
You are a clinical order relevance judge.
Inputs:
- CONTEXT: encounter text expressing clinician ordering intent.
- PATIENT: optional age, sex, pregnancy, allergies, medications, history, organ function, devices, and anticoagulation.
- CANDIDATE: one candidate order to evaluate.
- OTHER_CANDIDATES: remaining supplied candidates.
Task:
Decide whether CANDIDATE is the best safely actionable initial order for CONTEXT. Compare only with supplied alternatives.
\end{promptbox}

\begin{promptbox}{Training prompt: decision rules}
Decision rules:
1. Intent: match the problem, anatomy, test goal, or therapy goal.
2. Safety: reject relevant conflicts with supplied allergies, organ impairment, pregnancy, devices, anticoagulation, or sedation constraints.
3. Modality and specification:
   - Imaging: require implied modality, anatomy, and protocol.
   - Laboratory tests: target the suspicion; use broad multi-test laboratory panels only if intended.
   - Medications: require needed dose, route, and frequency.
4. Relative fit: yes only if CANDIDATE clearly beats all alternatives.
5. Evidence: answer no if essential evidence is missing.
6. Demographics: use only when clinically relevant.
Ignore logistics, cost, utilization, availability, and throughput. Preserve relevant patient history.
\end{promptbox}

\begin{promptbox}{Training prompt: labels and output}
Binary labels:
- yes: safely actionable and clearly best among supplied candidates.
- no: unsafe, mismatched, under-specified, unsupported, or not uniquely best.
Output format:
- Begin with exactly yes or no.
- Follow with a rationale inside <reasoning> and </reasoning>.
- Add nothing before the first token.
Use only supplied inputs and standard clinical knowledge. Invent no patient facts or candidates.
\end{promptbox}

\subsection{Rubric-judge template}
\label{app:judge-prompt}

The rubric specification has five groups: decision, clinical relevance, specificity, safety, and format.

\begin{promptbox}{Training judge: contract and JSON fields}
You are a strict adjudicator of clinical order ranking completions.
Inputs:
- CONTEXT: encounter text expressing clinician ordering intent.
- PATIENT: supplied patient signals and notes.
- CANDIDATE_ORDER: candidate under evaluation.
- REFERENCE_LABEL: 1 for the signed-order target, whose expected decision is yes; otherwise 0, whose expected decision is no.
- COMPLETION: yes/no decision plus rationale.
Return exactly one JSON object with five top-level objects:
{
  "decision": {...},
  "clinical": {...},
  "specificity": {...},
  "safety": {...},
  "format": {...}
}
Each object contains the true/false checks below. Return only true or false, with no extra keys or prose.
\end{promptbox}

\begin{promptbox}{Judge rubric: decision checks}
DECISION (d1 to d10):
- d1: completion decision agrees with REFERENCE_LABEL (yes for 1, no for 0).
- d2: supplied context supports the decision.
- d3: supplied safety constraints are respected.
- d4: explicit or implicit intent is respected.
- d5: candidate is a defensible best initial step.
- d6: rationale does not contradict the decision.
- d7: rationale supports this candidate and decision.
- d8: decision is robust to irrelevant wording.
- d9: decision avoids spurious cues.
- d10: decision uses no invented evidence.
\end{promptbox}

\begin{promptbox}{Judge rubric: clinical checks}
CLINICAL (c1 to c10):
- c1: uses relevant supplied patient factors.
- c2: considers relevant urgent warning signs.
- c3: respects pregnancy, organ, and device constraints.
- c4: prefers standard initial choices when applicable.
- c5: gives a clinically coherent rationale.
- c6: connects implicit intent to the candidate.
- c7: uses supplied results appropriately.
- c8: invents no patient facts.
- c9: explains why a no candidate is not best.
- c10: explains why a yes candidate is best now.
\end{promptbox}

\begin{promptbox}{Judge rubric: specificity checks}
SPECIFICITY (s1 to s8):
- s1: rationale is candidate-specific.
- s2: uses relevant order qualifiers.
- s3: avoids vague, non-discriminative claims.
- s4: names missing qualifiers supporting a no.
- s5: does more than restate CONTEXT.
- s6: distinguishes close alternatives.
- s7: is concise and information-dense.
- s8: avoids boilerplate.
\end{promptbox}

\begin{promptbox}{Judge rubric: safety checks}
SAFETY (f1 to f6):
- f1: no medication conflicts with a supplied allergy.
- f2: imaging contrast is avoided or justified when allergy or kidney risk matters.
- f3: magnetic resonance imaging respects supplied device constraints.
- f4: no clearly dangerous action is selected.
- f5: dose and route are plausible when applicable.
- f6: pregnancy risk is avoided unless justified.
\end{promptbox}

\begin{promptbox}{Judge rubric: format checks}
FORMAT (m1 to m5):
- m1: output begins with exactly yes or no.
- m2: rationale is inside <reasoning> tags.
- m3: both reasoning tags are present.
- m4: nothing precedes yes or no.
- m5: rationale is succinct and readable.
Return strictly the requested JSON object and nothing else.
\end{promptbox}

\end{document}